 \documentclass{vgtc}                 

\ifpdf
  \pdfoutput=1
  \usepackage{graphicx}
  \DeclareGraphicsExtensions{.pdf,.png,.jpg,.jpeg}
\else
  \usepackage{graphicx}
  \DeclareGraphicsExtensions{.eps}
\fi

\graphicspath{{figures/}{pictures/}{images/}{./}} 

\usepackage{microtype}                
\usepackage{textcomp}                 
\usepackage{mathptmx}                 
\usepackage{amsmath}                  
\usepackage{times}                    
\usepackage{cite}                     
\usepackage{url}                      

\usepackage{subcaption}               
\usepackage{placeins}                 
\usepackage{stfloats}                 
\usepackage{multicol}                 
\usepackage{booktabs}                 
\usepackage{tabu}                     
\usepackage{color}                    
\usepackage{enumitem}

\onlineid{0}

\vgtccategory{Research}

\title{Co-Located VR with Hybrid SLAM-based HMD Tracking\\ and Motion Capture Synchronization\thanks{Preprint. Accepted at GI VR/AR Workshop 2025 (Lecture Notes in Informatics).}}

\author{
  Carlos A. Pinheiro de Sousa\thanks{e-mail: carlos.pinheiro-de-sousa@uni-konstanz.de} \\
  \scriptsize University of Konstanz
  \and
  Niklas Gröne\thanks{e-mail: niklas.groene@uni-konstanz.de} \\
  \scriptsize University of Konstanz
  \and
  Mathias Günther\thanks{e-mail: mathias.guenther@uni-konstanz.de} \\
  \scriptsize University of Konstanz
  \and
  Oliver Deussen\thanks{e-mail: oliver.deussen@uni-konstanz.de} \\
  \scriptsize University of Konstanz
}

\teaser{
  \centering
  \includegraphics[width=0.95\linewidth]{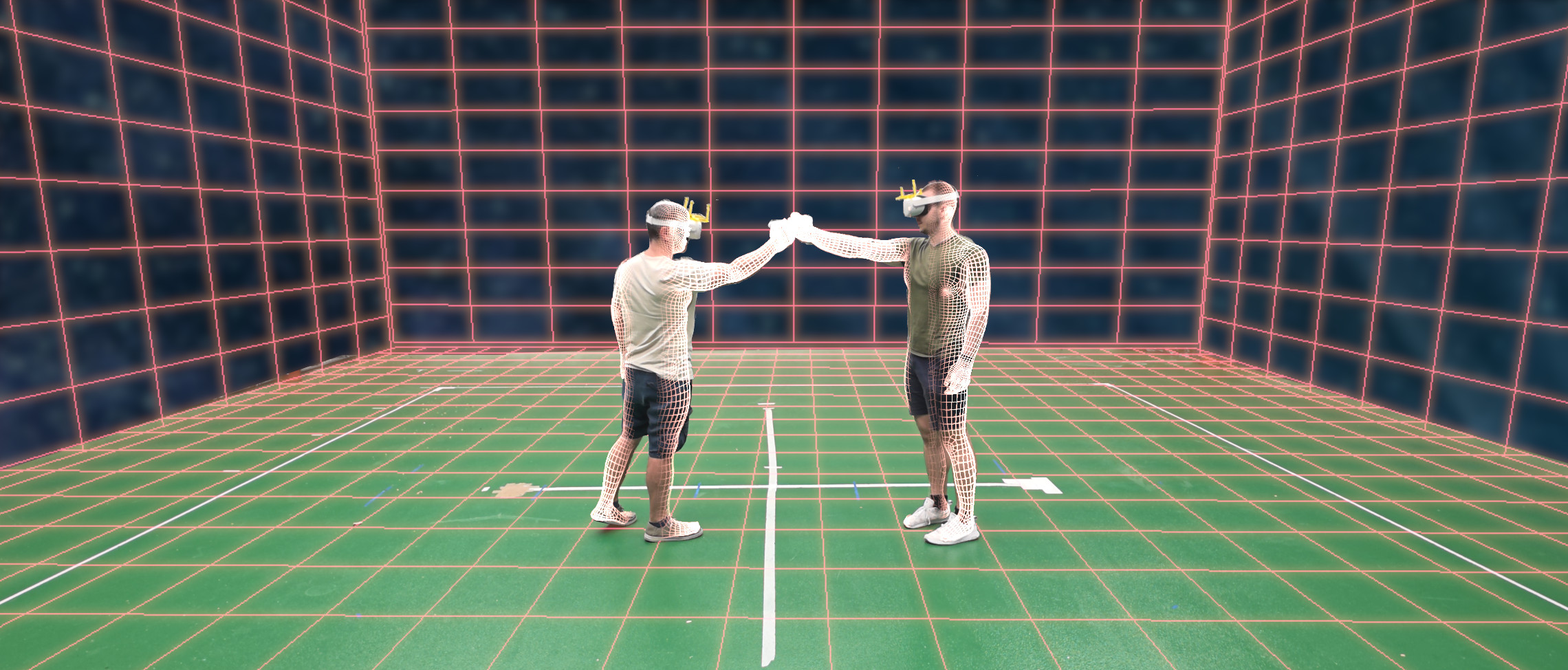}
  \caption{Co-located User Interaction: Fist Bump in VR.}
  \label{fig:cover}
}

\abstract{
We introduce a multi-user VR co-location framework that synchronizes users within a shared virtual environment aligned to physical space. Our approach combines a motion capture system  with SLAM-based inside-out tracking to deliver smooth, high-framerate, low-latency performance. Previous methods either rely on continuous external tracking, which introduces latency and jitter, or on one-time calibration, which cannot correct drift over time. In contrast, our approach combines the responsiveness of local HMD SLAM tracking with the flexibility to realign to an external source when needed. It also supports real-time pose sharing across devices, ensuring consistent spatial alignment and engagement between users. Our evaluation demonstrates that our framework achieves the spatial accuracy required for natural multi-user interaction while offering improved comfort, scalability, and robustness over existing co-located VR solutions.
}

\begin{document}
\maketitle

\keywords{Virtual reality, multi-user, motion capture, device tracking, synchronization}

\section{Introduction}
Virtual reality (VR) systems have evolved over decades, with immersion quality driven by hardware advancements and enhanced user interactions for richer perceptual experiences. Key specifications such as higher resolution, wider field-of-view, increased framerate, greater tracking accuracy, lower latency, amongst others are sought to deliver a user experience with strong \textit{presence}: the subjective feeling of being physically located within the virtual environment.

Multi-user VR enhances \textit{presence} through collaborative interactions in shared virtual spaces \cite{steuer1992defining,lombard1997heart}. Recent works define setups where users share both physical and virtual spaces as co-located VR \cite{podkosova2019walkable,defanti2019colocated, mcgill2020quest}. However, ensuring accurate and responsive tracking of users relative to each other remains a key challenge, as most off-the-shelf HMDs are not designed to provide co-located VR experiences.

Such tracking\footnote{Also referred commonly as \textit{localization}.}, typically describes the estimation of position and orientation of a device. On commercial HMDs tracking is achieved mostly through two approaches: 

\begin{enumerate}
    \item \textbf{Outside-in Tracking:} Relies on external base stations mounted in the physical environment. An example of such a system is the HTC Vive and their set of Lighthouse external base stations. These stations emit infra-red signal sweeps which are detected and processed by the headset and controllers for their localization estimation. This method is seen mostly on previous HMD generations and is being replaced by inside-out tracking.
    
    \item \textbf{Inside-out Tracking:} Is adopted by the majority of modern HMDs \cite{reimer2021colocation, collet2017slam}. This method employs Visual-Inertial Simultaneous Localization and Mapping (VI-SLAM), a technique widely used in robotics and XR, which relies on built-in cameras and additional inertial measurement units (IMUs). It provides six-degrees-of-freedom (6-DoF) localization information while utilizing a mapping of the surroundings. The main advantage of inside-out tracking system is that they provide a simpler setup and independent, self-contained tracking functionality. 
\end{enumerate}

However, despite several advantages of inside-out or SLAM-based tracking devices, these systems lack practicality for co-located VR. Each HMD creates an individual tracked map not inherently shared with other devices, and typically cannot detect other nearby devices, complicating co-location. As a result, even when users are physically side by side, they remain effectively invisible to one another in VR or are misaligned in virtual shared space with respect to their real world counterpart. 

To overcome the limitations of SLAM-based tracking in co-located VR, various system configurations have been explored as complementary external tracking and alignment solutions. The synchronization with such complementary external tracking devices is typically done either by:

\begin{enumerate}
    \item \textbf{One-time alignment\footnote{Although related literature typically refers to as one-time \textit{calibration}, in this work calibration describes a different process, therefore we keep the word for that purpose.}:} Which uses the external tracking information for setting up and initializing the system, relying thereafter on the individual HMD's  tracking capabilities.
    
    \item \textbf{Continuous alignment:} In this case, the external tracking system continuously overrides the HMD’s internal estimation, relying entirely on the external source for tracking throughout the session. 
\end{enumerate}

Among such external tracking systems are traditional motion capture (MoCap) which offer high-accuracy localization. However, when MoCap data is transmitted over networks, it introduces latency and jitter. Moreover, relying solely on MoCap for real-time, continuous head tracking often fails to meet the sensitive high demands such as low-latency and smooth tracking to avoid motion sickness. 

Adding to the complexity of such systems, SLAM-based HMDs are also not fail-proof: Tracking is disrupted by a number of known conditions such as image based artifacts and sensor imperfections which leads to drift accumulation and sometimes complete tracking break-down \cite{slam-handbook, bujanca2021robustslamsystemsyet, li2019survey, hu2024apple}.

Shared physical space multi-user VR applications, or co-located VR such as games, training, or collaborative design require precise, low-latency, and stable tracking to avoid cybersickness and maintain spatial consistency across users. 

These challenges highlight the need for a robust framework for co-location for multi-user VR systems that pertains both accuracy and critical low-latency smooth tracking. As such is the strong need to explore hybrid approaches that combine the strengths of existing systems while mitigating their limitations. 

\section{Related Work}
\label{sec:related-work}
Various approaches have been explored over the years to enable shared or co-located VR experiences, each presenting trade-offs between setup complexity and localization reliability. These include alignment methods such as manual alignment, image-based alignment techniques, and marker-based systems which includes motion capture. Given the context of this work, we also review collaborative SLAM approaches from both robotics and XR domains.

\subsection{Manual Alignment}
The most basic co-location methods rely on manual alignment, where users physically position and orient their HMDs at predefined reference points. One-point alignment \cite{mcgill2020quest} is simple but highly sensitive to angular error, causing significant spatial drift over distance. Two-point alignment \cite{mcgill2020quest} adds a second reference to reduce rotational ambiguity, improving accuracy but requires more setup and precise execution \cite{reimer2021colocation}. Both rely on one-time calibration and are impractical due to drift and alignment failure.

\subsection{Image-based Alignment}
Image methods use visual fiducials such as ArUco markers placed in the environment to establish a shared spatial reference. Each user’s device detects the marker and estimates its pose relative to it, aligning its virtual space accordingly.

The underlying method also known as \textit{perspective-n-point} estimates the camera pose, in this case HMD position and orientation, based on known 2D feature measurements from the markers and their respective projections into image plane \cite{opencvSolvePnP}. This is the underlying method used in \cite{reimer2021colocation}, although this improves accuracy and is less complex for setup compared to manual methods described previously, it is typically restricted to one-time alignment because of limited visibility of the marker during movement, making it also vulnerable to drift and tracking loss. 

Also in \cite{gugenheimer2016immersivedeck}, the method is used as inside-out continuous positional tracking by constantly tracking dozens of ceiling markers, which can be considered analogous to how modern HMDs estimate localization, but requires complex marker setup and users carrying a laptop. 

Finally, in another example, image markers are also used in \cite{defanti2019colocated}, however, in this case, they are used to track coordinate frame alignment between users instead of frame alignment with the physical environment. Yet, as noted by the author, the system is impractical due to low frame rate for detection and tracking of several users, noise corruption and further complications caused by natural occlusions by users movements.

Typically image-based alignment are better suited for short spatial range AR experiences, as many have been presented in the past. Because inherited noise from image sensors, along with limited field of view and occlusions this application has been limited in use in co-location VR. 

\subsection{Co-located XR and Collaborative SLAM}

Recent approaches to SLAM have increasingly focused on dense and high-fidelity mapping representation and scene reconstruction \cite{sandström2024splatslamgloballyoptimizedrgbonly, yan2024gsslamdensevisualslam}. However, such methods are impractical on resource-constrained platforms such as XR headsets or small mobile robots. In contrast, collaborative SLAM (C-SLAM) emphasize localization and relies on lightweight, efficient, and sparse methods, where multiple agents must operate in real-time with minimal latency and communication overhead \cite{slam-handbook, howard2006multi_robot, Lajoie_2024, bird2025dvmslamdecentralizedvisualmonocular}.

Co-located VR, where multiple users share a physical space and interact in a shared virtual environment, presents a problem closely related to collaborative SLAM: each XR device tracks independently using SLAM, however, no shared spatial reference exists by default. 

Despite this conceptual overlap, collaborative SLAM methods remain rare in XR, primarily due to hardware constraints and limited access to raw sensor data. Few works have adapted SLAM to XR; for instance, Mesh2SLAM \cite{desousa2025mesh2slamvrfastgeometrybased} operates on virtual meshes but is not intended for real-world sensor input, restricting its use to simulated environments.

Other approaches like ShareAR and SLAM-Share \cite{ran2019sharear,10.1145/3555050.3569142} support pose or map sharing in mobile AR via edge servers, but have not been extended to real-time multi-user VR on head-mounted displays mainly due to strict latency requirements, asymmetric tracking architectures, and platform restrictions.

Modern HMD devices like the Meta Quest use SLAM for standalone tracking, but each headset creates its own map without native sharing~\cite{reimer2021colocation,mcgill2020quest}. While Meta’s Shared Spatial Anchors~\cite{metaSSA2024} attempt to address this, they have seen limited adoption due to ecosystem lock-in, platform specificity, and complex integration. Moreover, such solutions are closed and opaque, offering little transparency or control for developers making them ill suited for scalable or research-oriented co-located VR systems.

\subsection{Motion Capture and Virtual Reality}

Motion capture (MoCap) is central to our proposed method, and its integration with virtual reality (VR) has been explored extensively in multi-modal systems. One of the earliest efforts, the VPL DataSuit from the late 1980s, enabled full-body tracking for immersive VR, though its adoption was limited by high cost and technical complexity~\cite{rheingold1991virtual, 10.1145/3555050.3569142}.

MetaSpace~\cite{Sra2015}, while not employing traditional IR marker-based MoCap, enabled co-located full-body VR using Kinect-based skeletal tracking mapped to shared virtual avatars. Although users could move naturally without controller markers and the cost benefit of using such RGB-D cameras, the system is constrained by limited range, narrow field of view, and unstable tracking.

Closer to our approach, \textit{Holojam}\cite{defanti2019colocated} used the Samsung GearVR, which was limited to 3-DoF -only orientation- combined with continuous motion capture positional tracking, enabling 6-DoF via an OptiTrack system. The author reported that network congestion introduced latency, which contributed to cybersickness in users. 

Similarly, Braun et al.\cite{braun2022virtual} proposed a firefighter training system using external continuous MoCap for positional head and body tracking combined with inside-out tracking for orientation. Unfortunately no latency measurements are available from this study.

Although motion capture systems can report latency under 10 ms \cite{qualisysLatency2024, viconLatency2024}, which might operate under ideal customized setups, under realistic and congested networks measured MoCap-related latencies range from 39 ms to over 100 ms~\cite{latoschik2017alpha,thomas2014evaluation,schmidt2018blended}, which can disrupt real-time interaction and immersion, contributing to cybersickness~\cite{defanti2019colocated, caserman2020, braun2022virtual}.

Caserman et al.~\cite{caserman2020} report that delays exceeding 63 ms can induce significant cybersickness symptoms. Van Damme et al.~\cite{app14062290} demonstrated that network latency in the range of hundreds of milliseconds negatively impacts task performance, mutual understanding, and perceived workload in collaborative VR scenarios. 

In contrast, modern HMDs offer fast and robust inside-out tracking with minimal latency and achieve 120 frames per second refresh rate, but they still lack practical shared reference frame methods for aligning multiple users in a common physical-virtual space, limiting their utility for co-located multi-user applications. These findings emphasize the need for latency-aware system design, particularly in multi-user environments like ours.

\section{Method}

We propose a novel, hybrid tracking approach with the best of both worlds, for an accurate yet smooth co-located multi-user VR experience: It combines the accuracy of motion capture with the high frame rate and low-latency tracking of SLAM-based HMDs to enable stable and responsive co-located VR. In this work, motion capture is used for occasional alignment of users in a shared physical-virtual space, after which each device relies on its own SLAM-based tracking for real-time, low-latency updates; concomitantly, realignment corrections are applied when necessary, ensuring both accuracy and scalability of the system.

Additionally, in order to provide consistent inter-user representations and to ensure smoother animations that enhance overall immersion, we directly transmit each user's position, including hand or controller data, to all other users, rather than relying on motion capture data. 

To the best of our knowledge, unlike prior approaches that rely either on continuous external tracking or one-time calibration without subsequent correction, our system mitigates jitter and latency that contribute to motion sickness, while maintaining alignment and enabling correction for gradual drift or complete tracking loss. In general terms, the concept of alignment consists in adjusting the provided tracking space from the device such that the virtual pose (position and orientation) point-of-view of the user corresponds to that of the motion capture user pose.   

The setup was developed with Qualisys motion capture system and Quest 2 HMDs, however its modular design allows any motion capture or other external tracking system and different HMDs to be used.  Resources are publicly available on Github \footnote{https://github.com/niklas-groene/Co-Located-VR}.

\subsection{System Overview}

Here, the processes of the framework are enumerated with a brief description; each is then described in detail in the following subsections.

\begin{enumerate}
    \item \textbf{Extrinsics Calibration}~(Section~\ref{sec:Extrinsics Calibration}): The process for initial calibration procedure where the device’s tracking pivot (eye center) with respect to the mounted mocap frame is estimated. 

    \item \textbf{Virtual Camera Alignment}~(Section~\ref{sec:Virtual Camera Alignment}): The alignment of the tracking space with MoCap physical space for co-location. 
    
    \item \textbf{Initialization}~(Section~\ref{sec:Initialization}): validates input data from motion capture and HMD tracking to compute initial alignment ~(Section~\ref{sec:Virtual Camera Alignment}) before starting Tracking Main Loop ~(Section~\ref{sec:Tracking Main Loop}). 
    
    \item \textbf{Tracking Main Loop}~(Section~\ref{sec:Tracking Main Loop}): After initialization the system continues with HMD device tracking, while continuously monitoring drift for occasional corrections.
    
    \item \textbf{Dynamic Alignment Correction}~(Section~\ref{sec:Dynamic Alignment COrrection}): Corrects drift by re-alignment during runtime.
\end{enumerate}

\subsection{Extrinsics Calibration}
\label{sec:Extrinsics Calibration}
In this section, one of the essential steps required for achieving co-location alignment between two distinct tracking systems: external motion capture and internal HMD inside-out tracking is described. Accurate spatial alignment demands a pre-calibration phase that determines extrinscis offset between the aforementioned coordinate frames.

\subsubsection{Motion capture and Eye Center Offset}

\begin{figure}[ht]
    \centering
    \includegraphics[width=0.2\textwidth]{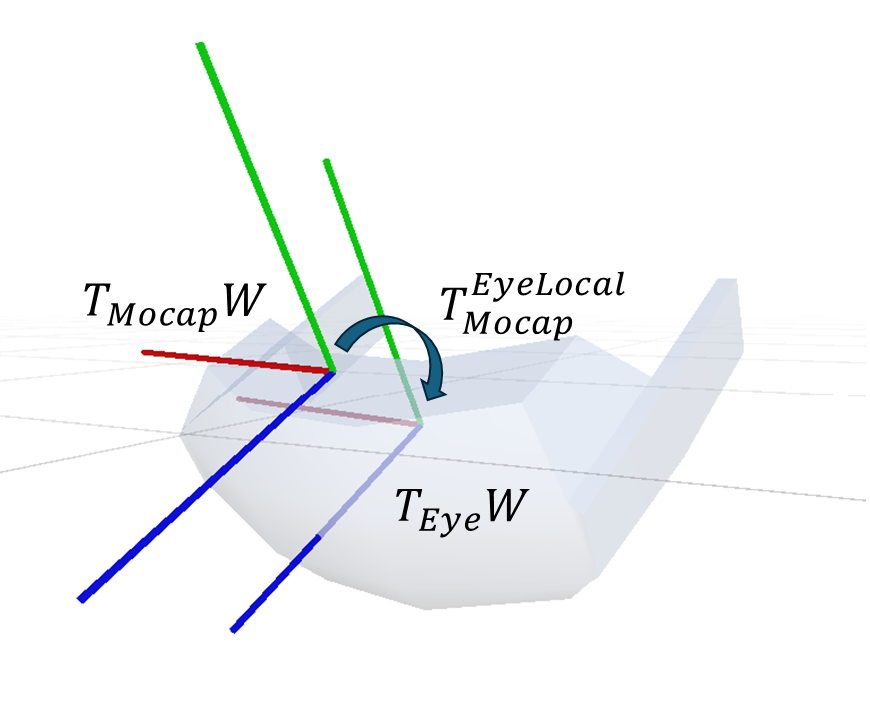}
    \caption{Rigid pose offset $T^{\mathrm{EyeLocal}}_{\mathrm{Mocap}}$      (translation and rotation)  from motion capture frame $T_{\mathrm{Mocap}}W$ to eye center frame $T_{\mathrm{Eye}}W$}
    \label{fig:XROrigin_01}
\end{figure}

The placement of reflective markers for motion capture defines a coordinate frame with position and orientation. Ideally it would coincide with the HMD’s internal tracking frame, however, perfect alignment between this external MoCap frame and the device's internal eye-centered tracking pivot is not possible. Since the MoCap rigid body is mounted externally, it introduces a fixed  offset (translation and rotation), from the so-called eye center frame, which is the pivot frame used for internal system tracking. This relationship is illustrated in Figure~\ref{fig:XROrigin_01}.

Additionally, most commercial HMDs do not specify the exact location of the eye center, making accurate alignment difficult. Typically, this point is described merely as the midpoint between the user’s eyes~\cite{optitrack_unity_hmd_setup}. However, inaccurate eye center pose estimation can significantly degrade tracking consistency.

Next, the rigid-body transform from the MoCap frame to the eye center estimation is described.

\subsubsection{Eye Center Frame Estimation}

We estimate the rigid transformation \( T^{\mathrm{EyeLocal}}_{\mathrm{Mocap}} \in \mathrm{SE}(3) \), which encodes the constant spatial offset from the MoCap body frame to the HMD’s eye center frame and posteriorly include such computation for the alignment procedure. 

Although heuristic estimates of such offset can be made, if good quality motion capture streaming is available, optimization processes can provide accurate offset estimates. The optimization processes available data which consists of world-referenced poses from the MoCap system \( T_{\mathrm{Mocap}}W(t) \), and from the HMD's eye-centered tracking system \( T_{\mathrm{Eye}}W(t) \), where the latter typically represents the virtual camera’s pose.

The procedure begins by aligning the trajectories of both frames using a similarity transformation~\cite{umeyama1991least}. This can be further refined for the fixed transform \( T^{\mathrm{EyeLocal}}_{\mathrm{Mocap}} \) by minimizing the following objective:
\[
\min_{T^{\mathrm{EyeLocal}}_{\mathrm{Mocap}}} \sum_t \left\| T_{\mathrm{Eye}}W(t) - T_{\mathrm{Mocap}}W(t) \cdot T^{\mathrm{EyeLocal}}_{\mathrm{Mocap}} \right\|^2
\]

Such estimate of the eye center would correspond to the virtual camera model used for rendering in the corresponding game engine \footnote{Although in VR two cameras are used, one for each eye and are typically offset from such center point.}. Next, a description of such alignment implementation for the framework is presented.


\subsection{Virtual Camera Alignment}
\label{sec:Virtual Camera Alignment}
The OpenXR API~\cite{openxr_tutorial_graphics} provides an interface to real-time HMD tracking data. In this system,  Unity 3D is used with the XR Interaction Toolkit, though the method generalizes to any engine that supports the same interface standard~\cite{unity_openxr_input}.

\begin{figure}[ht]
    \centering
    \includegraphics[width=0.3\textwidth]{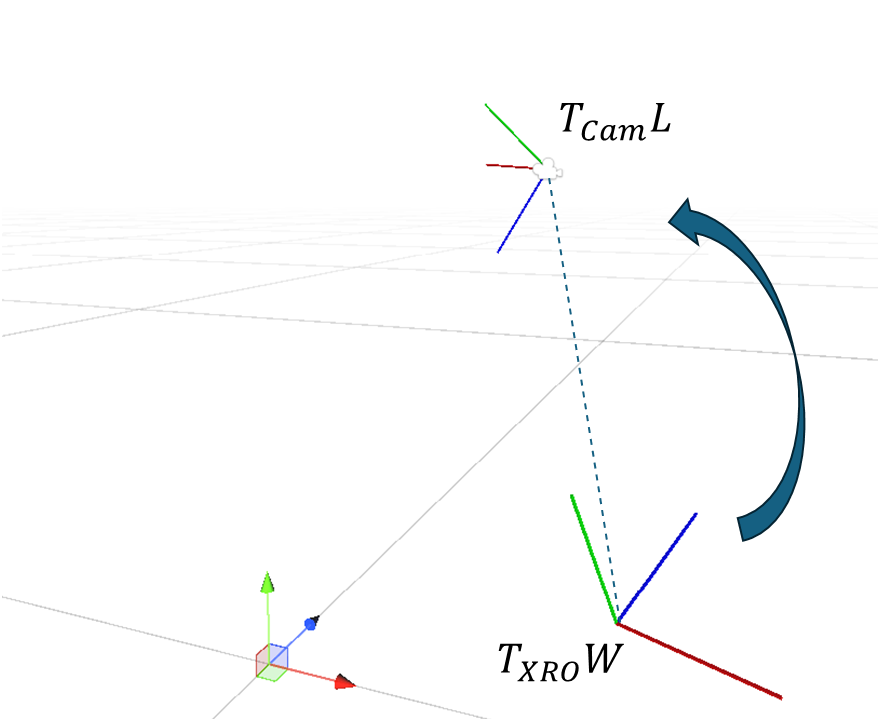}
    \caption{The virtual camera pose $T_{\mathrm{Cam}}L$ with respect to its parent, the XR Origin $T_{\mathrm{XRO}}W$.}
    \label{fig:XROrigin_02}
\end{figure}

In Unity, the virtual camera local pose is defined relative to an XR Origin world transform, which we abreviate as $T_{\mathrm{Cam}}L$ and $T_{\mathrm{XRO}}W$ respectively. This hierarchy is shown in Figure~\ref{fig:XROrigin_02}.

\begin{figure}[!htbp]
  \centering
  \includegraphics[width=0.3\textwidth]{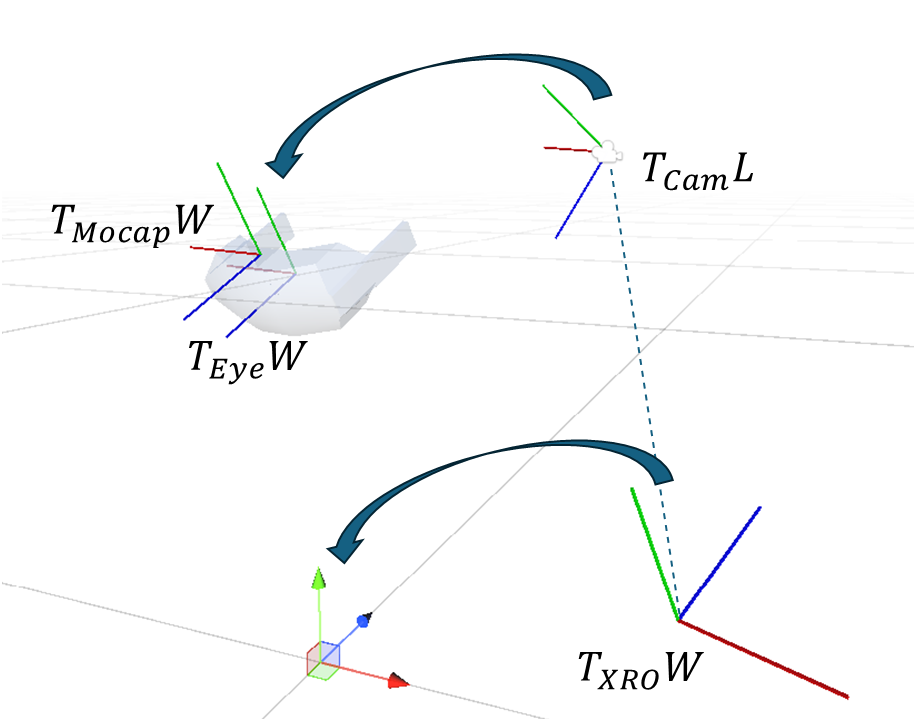}
  \caption{Co-location alignment: determine $T_{\mathrm{XRO}}W$ such that $T_{\mathrm{Cam}}L$ = $T_{\mathrm{Eye}}W$.}
  \label{fig:XROrigin_03}
\end{figure}

To correctly align the virtual camera with the calibrated eye center pose, it is necessary to adjust the XR Origin transform since directly manipulating virtual camera is not possible: The alignment procedure computes the correct XR Origin transform such that the virtual camera coincides with the estimated eye center. This concept is illustrated in Figure~\ref{fig:XROrigin_03}, a formal definition follows.

Let's assume the following definitions:

\begin{itemize}[itemsep=0pt, topsep=1pt, leftmargin=*]
 \item \( T_{\mathrm{Mocap}}W \in \mathrm{SE}(3) \): MoCap pose in world coordinates
  \item \( T^{\mathrm{EyeLocal}}_{\mathrm{Mocap}} \in \mathrm{SE}(3) \): fixed transform from MoCap to eye center
  \item \( T_{\mathrm{Cam}}L \in \mathrm{SE}(3) \): Virtual camera pose relative to XR Origin
  \item \( T_{\mathrm{Eye}}W \in \mathrm{SE}(3) \): eye center pose in world coordinates
  \item \( T_{\mathrm{XRO}}W \in \mathrm{SE}(3) \): XR Origin pose in world coordinates (to be computed)
\end{itemize}

The world pose of the eye center is computed utilizing the previously obtained calibration offset from the mocap frame:
\[
T_{\mathrm{Eye}}W = T_{\mathrm{Mocap}}W \cdot T^{\mathrm{EyeLocal}}_{\mathrm{Mocap}}
\]

To ensure the virtual camera coincides with this world pose, XR Origin must be set such that:
\[
T_{\mathrm{Eye}}W = T_{\mathrm{XRO}}W \cdot T_{\mathrm{Cam}}L
\]

Solving for the XR Origin’s world transform:
\[
T_{\mathrm{XRO}}W = T_{\mathrm{Eye}}W \cdot \left(T_{\mathrm{Cam}}L\right)^{-1}
\]

\subsubsection{Solving for Tilted tracking space}

Although the mathematical formulation solves for the alignment at first glance, one additional practical aspect must be addressed: the tilting of the tracking space. 

While manually setting the XR Origin does align the virtual camera with the HMD’s eye center in terms of 6-DoF pose, this adjustment is artificial and also implicitly reorients the entire tracking coordinate frame; In simple terms, if the user is looking downward during alignment, the tracking space will inherit this orientation. As a result,  subsequent tracking will be biased: Even though the alignment initially appears correct, when the user walks forward in physical space, their virtual movement will follow the tilted frame, causing them to walk toward the virtual floor. To address this issue, a reduction of the alignment from full 6-DoF considers only position and yaw, as similarly done in \cite{Razzaque2001RedirectedWalking}. This way it is possible to maintain correct floor leveling of the tracking space.

With the provided description of the core method, Extrinsics Calibration and Alignment, we next clarify other processes of our framework: starting with Initialization.

\subsection{Initialization}
\label{sec:Initialization}

In a shared physical area, $N$ users are tracked by motion capture system. Each user's HMD is equipped with retro-reflective markers, enabling tracking of their 6-DoF pose data. 

\begin{figure}[h]
\centerline{\includegraphics[width=1\linewidth]{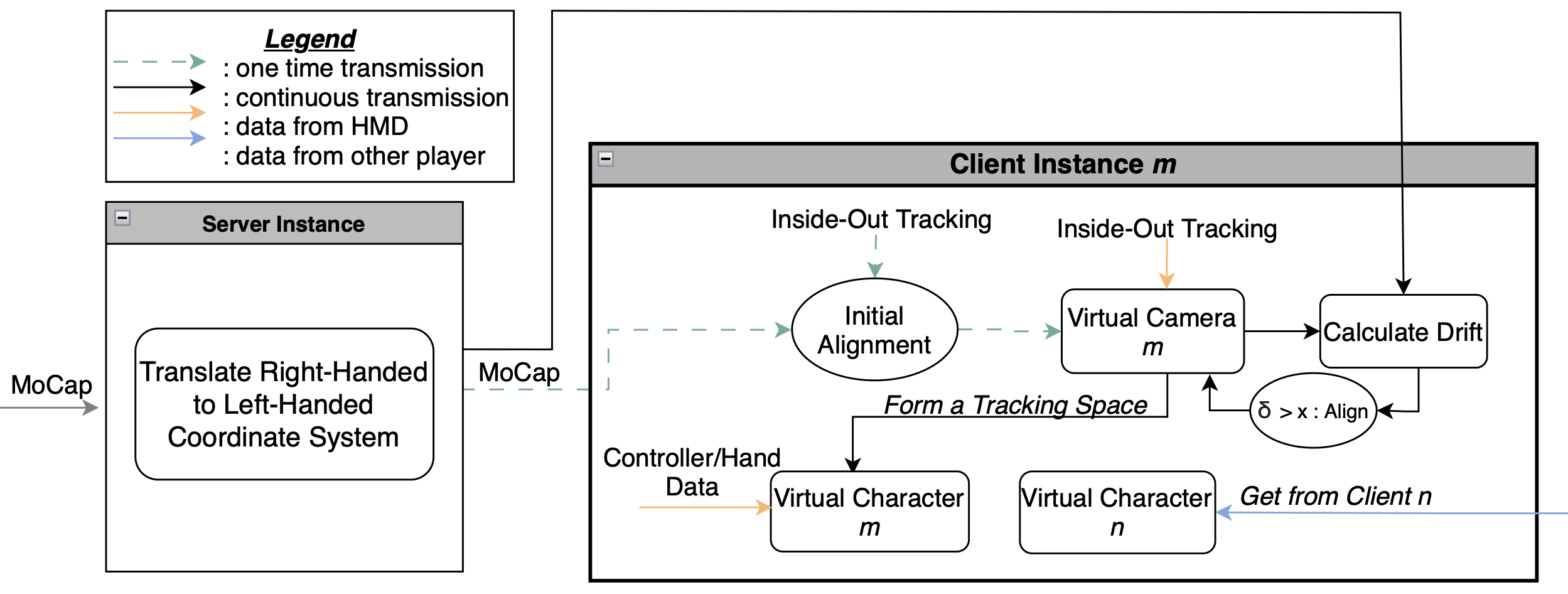}}
\caption{Schema of the proposed framework with one exemplary user-instance.}
\vspace{-0.1cm}
\label{fig:phys}
\end{figure}

The pose information is packaged into data packets and streamed in real-time from the Qualisys Track Manager (QTM) software to a server instance running the Unity 3D engine. The processed data is being transmitted via TCP/IP, adhering to the standard QTM RT server protocol version 1.25. Upon receipt, the pose data are processed and transformed to match Unity’s coordinate system convention, which requires conversion from a right-handed to a left-handed system, before being distributed to the corresponding user instances (see Fig.~\ref{fig:phys}).

During initialization, each user instance compares incoming server packets with predefined placeholder values. While packets match these values, it indicates that motion capture data is not yet being transmitted. Once a packet differs, the system assumes the values originate from motion capture input, switches to tracking mode, and ends the verification step.

\subsection{Tracking }
\label{sec:Tracking Main Loop}
After initialization and validation of HMD tracking and motion captured data, a state change to Tracking  will perform an initial alignment and continue tracking upon the adjusted tracking space.

For effective multi-user co-localization, each client instance also acts as a server, transmitting the pose data of its shared game objects to all other clients. This is depicted in figure \ref{fig:phys} as \textit{client m} receiving data from \textit{ client n}. The functionality is implemented using the Colibri framework \cite{hubenschmid2023}, which provides the underlying server infrastructure.

We opted for minimalist character representation, only user head and hands pose information are sent/received and are represented as geometric primitives, however such framework can be extended to support other types of data distribution, for example full-body pose estimation. Consequently, it becomes possible to share virtual full-body characters realistically animated, representing their real-world counterpart across the network. However, achieving fluid motion and managing this extensive data transmission is beyond the scope of this work.

\subsection{Dynamic Alignment Correction}
\label{sec:Dynamic Alignment COrrection}
Discrepancies between inside-out tracking and the external motion capture tracking causes co-location misalignment. In order to mitigate such issues, this framework performs dynamic alignment correction. As a result, such dynamic alignment places our framework as a hybrid system within the established synchronization category of either one-time calibration and continuous external tracking.  

The residual tolerance defines how fine grained alignment should be maintained with external tracking and is exposed as configurable parameter for ease of use. Threshold parameters allows system's alignment precision with external tracking to be managed: Defining an acceptable residual threshold, which does not hinder users' VR experience/immersion and determining how corrections are applied, for example smooth transitions should account for human perceptual factors not covered in this work. Yet, because modern HMDs typically provide high-quality tracking performance \cite{holzwarth2021comparing,hu2024apple}, assuming no failure on the part of the external tracking system, i.e. motion capture, re-alignment corrections are expected to be less frequent.

Next a description of how the framework has been evaluated and following, the results with demonstration of measurements is presented.

\section{Evaluation}
\label{sec:Evaluation}
\begin{figure}[h]
\centerline{\includegraphics[width=0.75\linewidth]{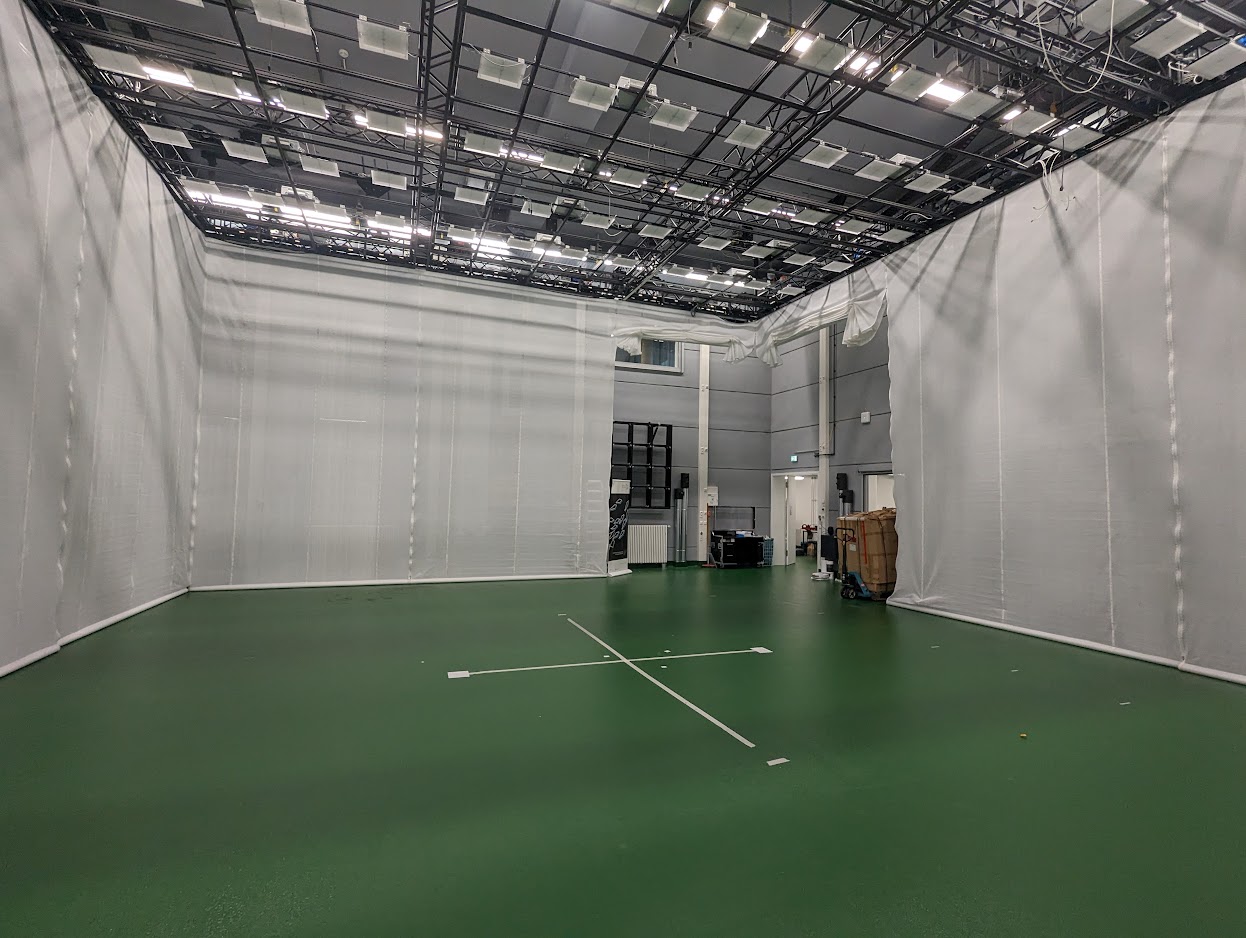}}
\caption{Environment used for testing.}

\label{fig:IH}
\end{figure}

A diverse set of evaluation motions has been selected which include different walkable circuits for single-user and multi-user interaction specifically intended for co-location reliability.  The physical space involved a room equiped with an array of 38 Arqus A12 cameras and an available area of 7m x 7m, see Figure \ref{fig:IH}.

Latency was also measured, not only to highlight its critical role co-location in frameworks, but also because latency will influence the measured tracking accuracy stemming from computed errors with outdated measurements from motion capture. 

\subsection{Singe-user Evaluation}

For single-user evaluation, we are mainly interested in determining the accuracy of the alignment by trajectory error verification through \textit{Absolute Trajectory Error} (ATE) which is expressed as \textit{RMSE}. 

\begin{figure}[h]
\centering
\includegraphics[width=0.75\linewidth]{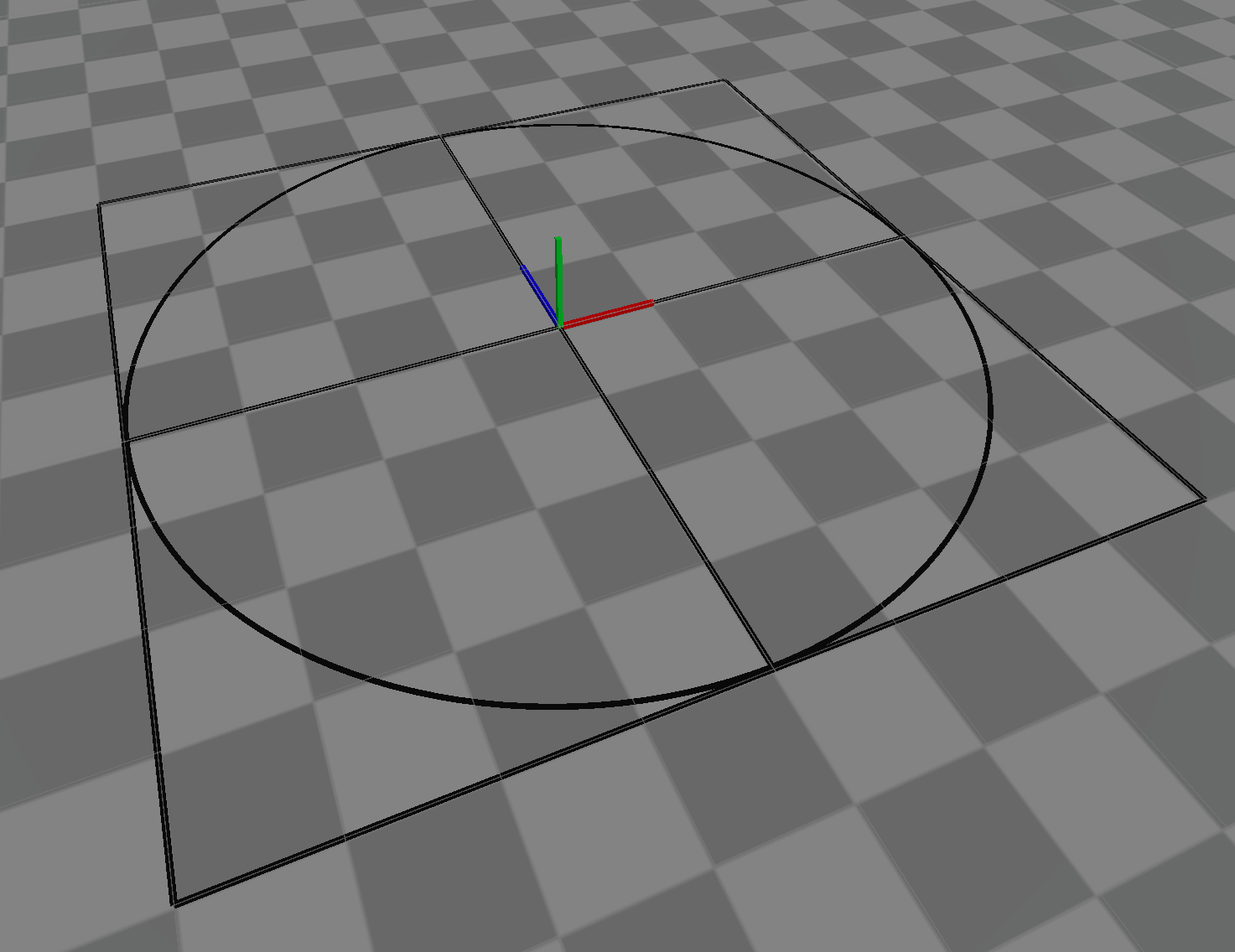}
\caption{Motion circuit, used for guiding users.}
\label{fig:circuit}
\end{figure}

Inspired by works performed by \cite{holzwarth2021comparing, hu2024apple}, we opted to create three distinct walk motions: The user is guided to perform specific walk motion on a circuit placed in the virtual environment, see figure \ref{fig:circuit}.

\begin{enumerate}
    \item \textbf{Straight Line Motion:}
    The user walks a straight line back and forth repeatedly. 
    
    \item \textbf{Scanning Motion:}
    The user walks in a circle form while facing the center of the scene.
    
    \item \textbf{Patrol Motion:}
    The user walks a quadrangular motion while facing the direction of movement.
\end{enumerate}

For each of the different motion experiment repetition, all execution is done in a single run 10 times, without restarting or realigning the system, with the assumption to better observe cumulative or gradual drift.

\subsection{Multi-user Evaluation}
The multi-user evaluation was performed in three independent runs. In each run, both users followed distinct motion paths around the circuit with translations and rotations (up to 90°), then walked toward the center and carried out a high-five or fist-bump interaction. Afterwards, they resumed the motion paths and repeated the interaction, completing this pattern four times per run.

\begin{figure}[h]
\centering
\includegraphics[width=0.75\linewidth]{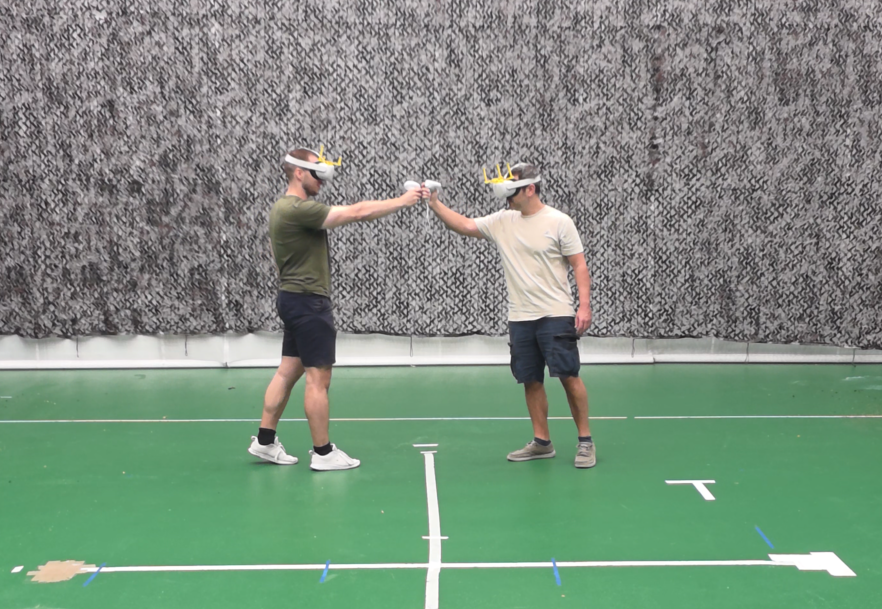}
\caption{Fist bump interaction, used for co-location validation.}
\label{fig:FistBumpPicture}
\end{figure}

The experiment design addresses key challenges in co-located VR through the following elements:

\begin{enumerate}
    \item \textbf{Translations and large rotations:}
    Certain motions are known to degrade tracking accuracy \cite{mcgill2020quest, hu2024apple}. Errors in center-eye offset or calibration become  apparent during: 1) translations, for example with positional drift due to ill-aligned yaw values; and 2) large rotations, due to shifts in the effective pivot point -which offsets the eye-center from its actual true position.
    
    \item \textbf{Fist bump interaction:}
    Performing a fist bump after both translation and rotation requires precise spatial alignment between users—not just visually, but in terms of tracked transform chain dependencies part of the VR rig, namely the controllers. 

    \item \textbf{Repetition:}
     To confirm all of the above, the experiment is performed repeatedly as users continuously change their global orientation and position in space verifying that the alignment remains consistent throughout.
     
\end{enumerate}
   
   Next we present the results from the described evaluations.

\section{Results}

\subsection{Latency}

The latency between the motion capture system and the virtual camera tracking was estimated via cross-correlation of signals \footnote{Given the Qualisys motion capture system's lack of Round-Trip Time (RTT) measurements and reliable timestamp synchronization for Unity}, which yields the temporal offset (in frames) between the two signals.  A sample of the latency between signals can be observed in figure \ref{fig:latency}.

\begin{center}
\includegraphics[width=1\linewidth]{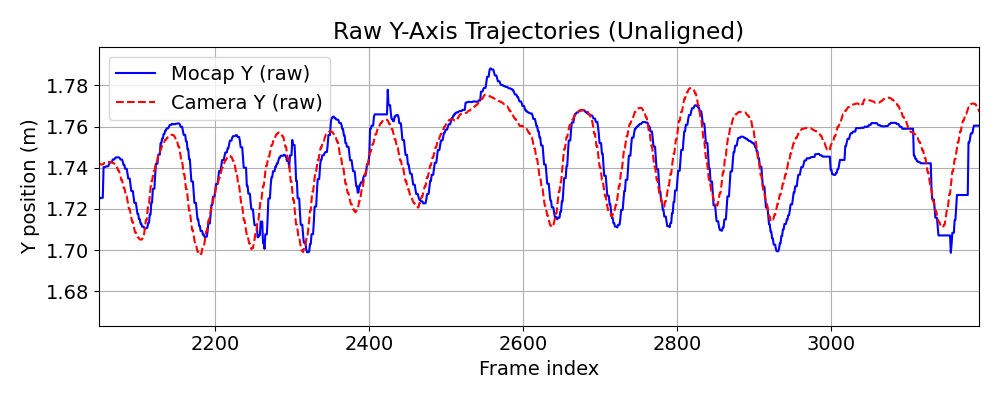}
\captionof{figure}{Latency across motion types.}
\label{fig:latency}
\end{center}
The data was logged in Unity using \textit{FixedUpdate()} at a fixed timestep of 0.01 s (100 Hz). Across all test sequences, the shift was consistently 7 frames, which corresponds to a latency of approximately 70 ms. Although frame-based estimation limits temporal resolution and may not capture real-time network variability, these constraints are negligible in this setup due to the stable logging frequency and consistent signal alignment observed.

 This confirmed latency further supports the benefit of our approach, which reinforces the case for leveraging on-device inside-out tracking for co-location instead of continuous pose streaming, ensuring that the responsiveness required for VR experience is not compromised.

\subsection{Single-user Results}

The accuracy was evaluated during single-user experiments designed to assess positional tracking performance under the controlled motion patterns. The \textit{ATE} with results can be found in table \ref{tab:singleATETable}.





\begin{table}[h]
\centering
\caption{Average ATE RMSE for Single-User Motion}
\begin{tabular}{lccc}
\hline
Motion & ATE & Unit \\
\hline
Line   & 3.410 & cm \\
Patrol & 3.102 & cm \\
Circle & 4.880 & cm \\

\hline
\end{tabular}
\label{tab:singleATETable}
\end{table}

The average ATE observed over 10 repetitions per walk experiment reflect nominal ranges according to their inherent motion complexity \cite{boulo2023validity,hu2024apple},  involving frequent rapid turns. Such motions are known to challenge SLAM-based tracking and may expose sub-optimal calibration of the eye-center offset. 

This is consistent when considering the motions and the values shown in Table~\ref{tab:singleATETable}: as Patrol motion has the lowest error, while Line, despite being the simplest straight-line translation motion, involves repeated 180° turns, and Circle requires the user to continuously reorient to keep the scene center in view, both introducing frequent rapid viewpoint shifts that increase tracking difficulty.

\begin{figure*}[t!]
    \centering
    \includegraphics[width=0.32\textwidth]{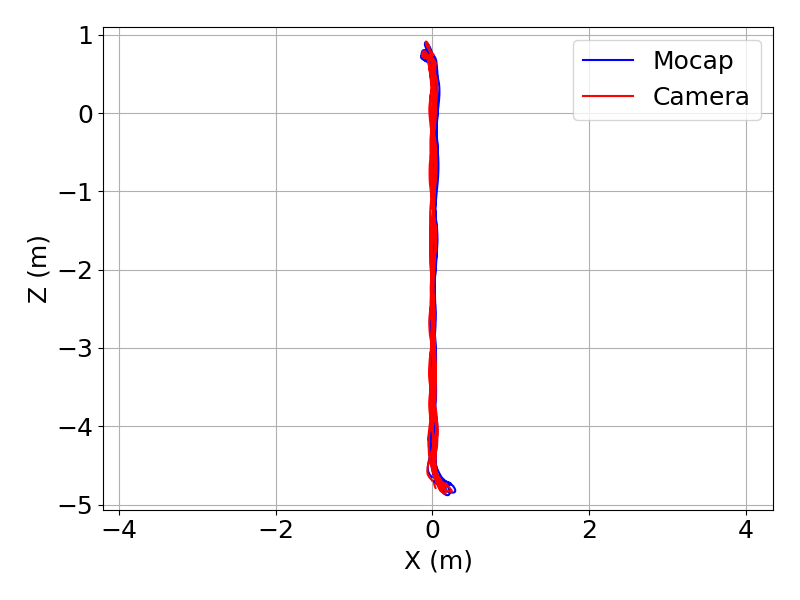}
    \includegraphics[width=0.32\textwidth]{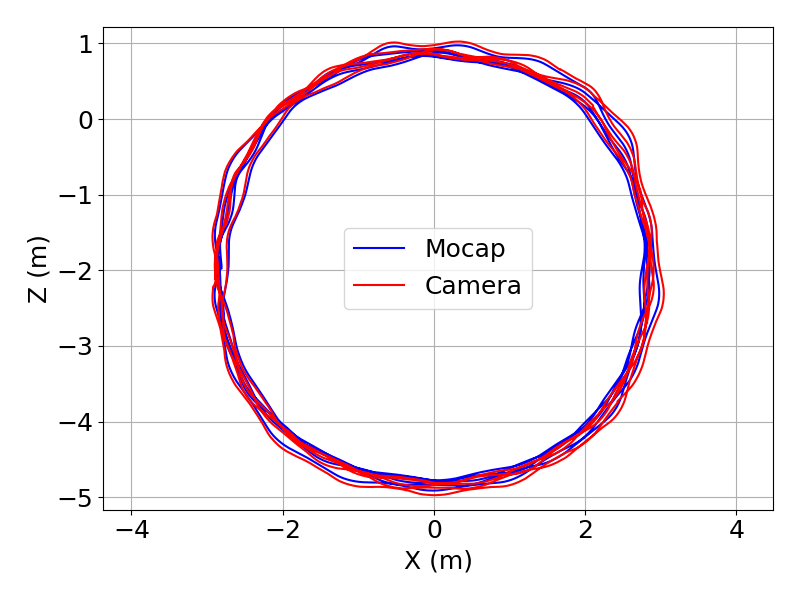}
    \includegraphics[width=0.32\textwidth]{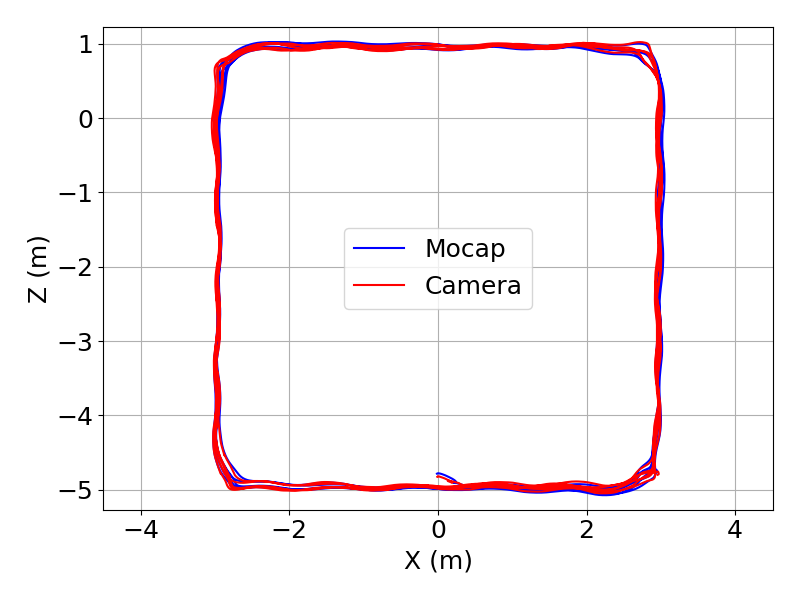}
    \caption{Top-down views of single-user motion patterns. Left: line motion. Middle: circle. Right: patrol.}
    \vspace{-0.55cm}
    \label{fig:movementPatterns}
\end{figure*}

Top-view trajectory plots shown in Figure~\ref{fig:movementPatterns} illustrate the consistent alignment throughout the different movement patterns.

\subsection{Multi-user Results}

As in the single-user evaluation, tracking remained stable despite complex, random, and rapid motions involving multiple users. The higher average ATEs are expected, given complex motion and interaction between users, nevertheless, as shown in Table~\ref{tab:multiATETable}, the ATE values remain well below the most difficult tracking scenario errors of approximately 12 cm reported in \cite{hu2024apple}.

\begin{table}[h]
\centering
\caption{Average ATE RMSE for Multi-User Motion and Interaction}
\begin{tabular}{lccc}
\hline
Motion & ATE & Unit \\
\hline
User 1  & 5.192 & cm \\
User 2  & 5.184 & cm \\

\hline
\end{tabular}
\label{tab:multiATETable}
\end{table}







\begin{figure*}[t!]
    \centering

    \begin{minipage}[t]{0.248\textwidth}
        \includegraphics[width=\linewidth]{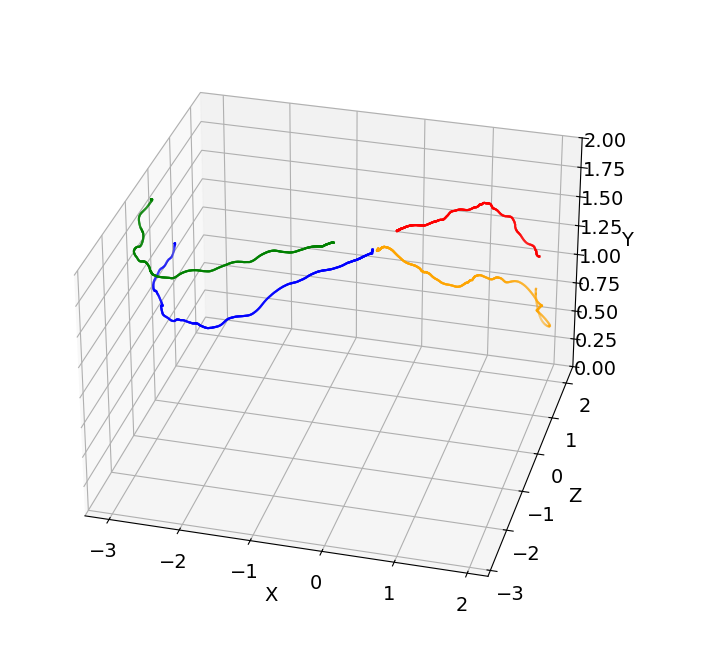}
        \caption*{Fist bump 1 (3D)}
    \end{minipage}\hfill
    \begin{minipage}[t]{0.248\textwidth}
        \includegraphics[width=\linewidth]{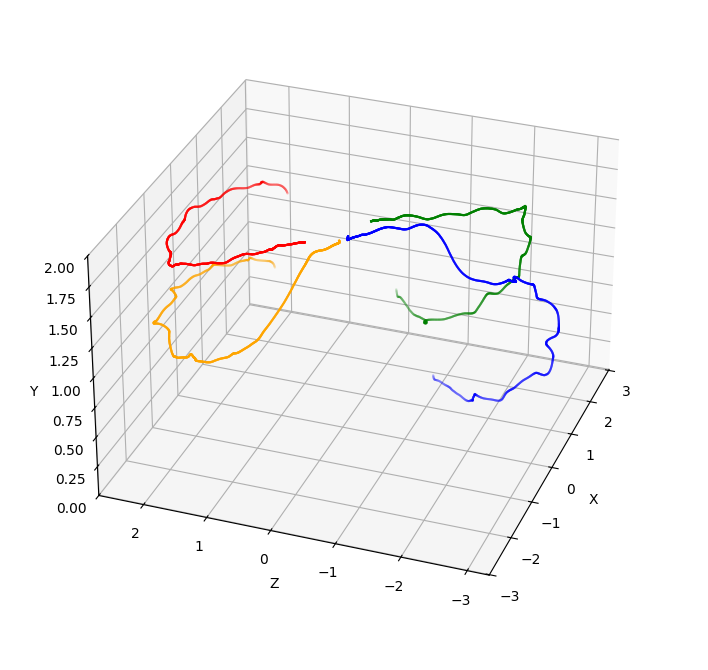}
        \caption*{Fist bump 2 (3D)}
    \end{minipage}\hfill
    \begin{minipage}[t]{0.248\textwidth}
        \includegraphics[width=\linewidth]{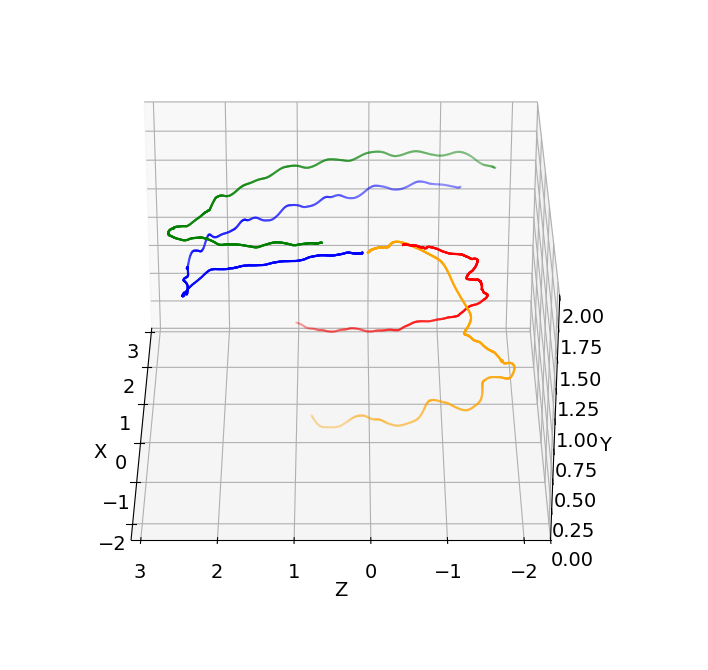}
        \caption*{Fist bump 3 (3D)}
    \end{minipage}\hfill
    \begin{minipage}[t]{0.248\textwidth}
        \includegraphics[width=\linewidth]{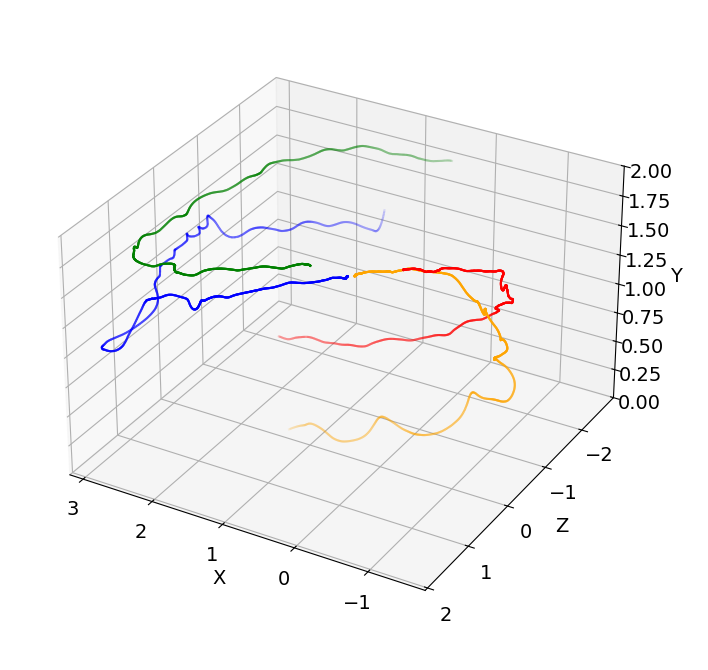}
        \caption*{Fist bump 4 (3D)}
    \end{minipage}

    \vspace{0.2em}

    \includegraphics[width=0.25\textwidth]{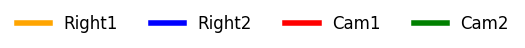}

    \vspace{0.5em}

    \begin{minipage}[t]{0.248\textwidth}
        \includegraphics[width=\linewidth]{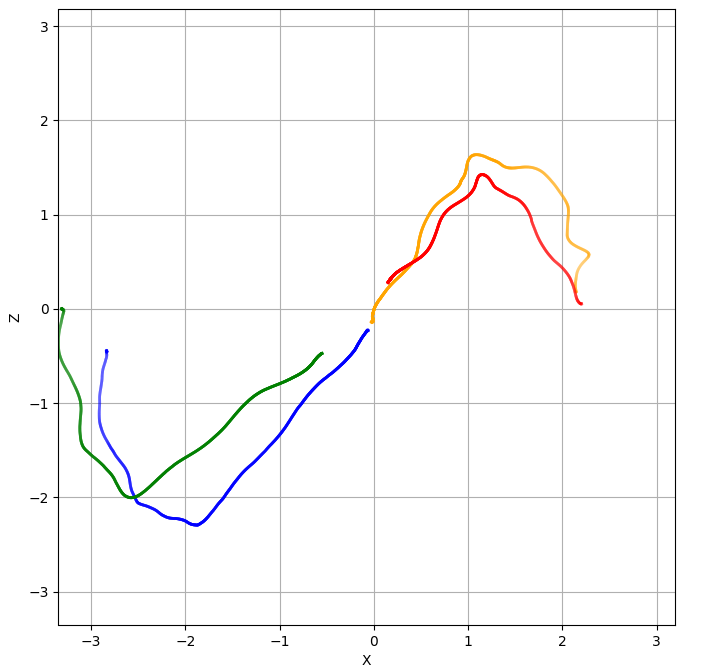}
        \caption*{Fist bump 1 (Top-down)}
    \end{minipage}\hfill
    \begin{minipage}[t]{0.248\textwidth}
        \includegraphics[width=\linewidth]{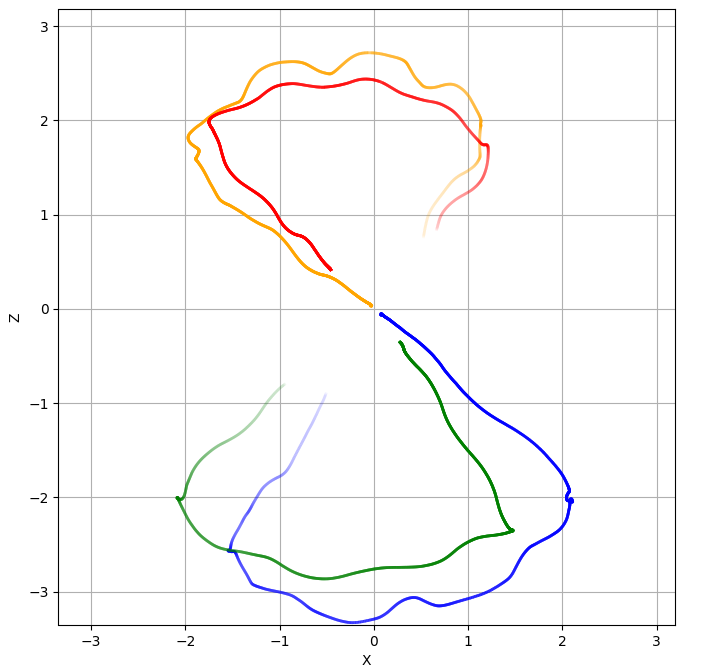}
        \caption*{Fist bump 2 (Top-down)}
    \end{minipage}\hfill
    \begin{minipage}[t]{0.248\textwidth}
        \includegraphics[width=\linewidth]{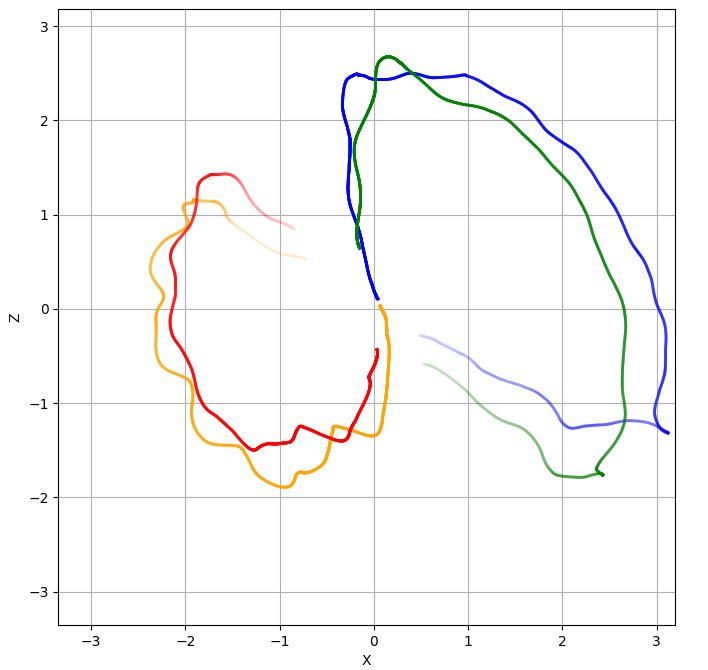}
        \caption*{Fist bump 3 (Top-down)}
    \end{minipage}\hfill
    \begin{minipage}[t]{0.248\textwidth}
        \includegraphics[width=\linewidth]{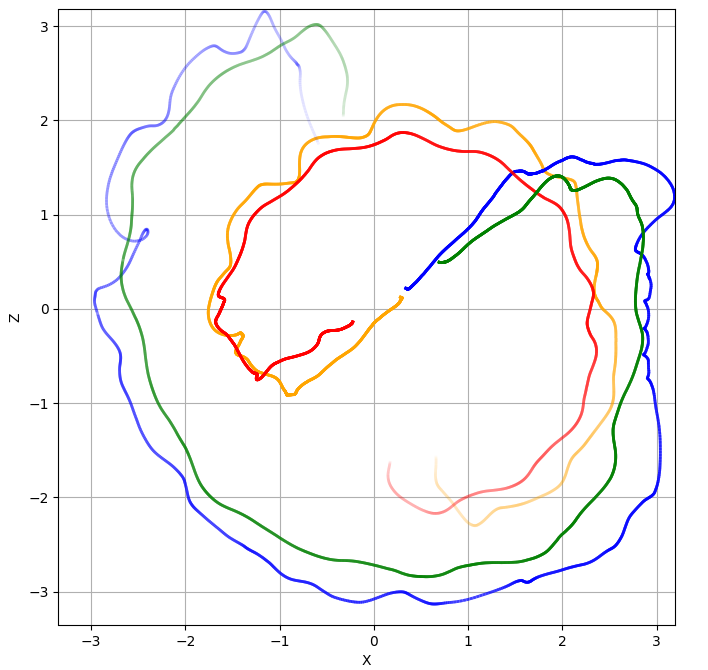}
        \caption*{Fist bump 4 (Top-down)}
    \end{minipage}

    \caption{Fist bump events showing 3D trajectories (top row), legend (center) where \textit{cam1} and \textit{cam2} correspond to the users 1 and 2 virtual cameras and \textit{right1} and \textit{right2} the user's right controller, and top-down views (bottom row) for four selected frames.}
    \vspace{-0.5cm}
    \label{fig:fistbump_layout}
\end{figure*}

The reported values are averages over three experiments. In each experiment, participants completed four uninterrupted cycles: they walked to the center, performed a fist-bump, resumed walking, and repeated the sequence. All fist-bump trials (Fig.~\ref{fig:fistbump_layout}) were successfully completed; participants consistently aligned and executed the interaction without perceptible misalignment.

These results demonstrate the system’s ability to support dynamic, multi-user co-located interactions with high spatial consistency. Despite the modest number of repetitions, the trials provide empirical evidence of accurate, stable co-location alignment during dynamic interactions, even in the presence of latency and motion-induced noise.

\vspace{-1em}

\section{Discussion}

The evaluation demonstrates that the hybrid tracking approach achieves robust and accurate co-location in VR, maintaining spatial alignment even under varied and dynamic motion. The evaluations and results empirically show that this precision supports close-range interactions such as fist bumps and collaborative manipulation, where consistency is critical for presence and usability.

Although motion complexity slightly increased ATEs, the system remained stable and repeatable. Errors likely stem from rapid movements and imperfect calibration offsets. A key strength is the minimal reliance on continuous external tracking, instead operating primarily on smooth HMD-SLAM pose estimation, thereby avoiding the latency and jitter that often cause motion sickness in externally aligned systems.

Future work includes scaling the hybrid method to larger environments and more users to assess robustness under greater spatial and social complexity. The current reliance on motion capture also underscores the need to explore alternatives, particularly SLAM-based systems with cross-device map sharing and occasional alignment. With modern HMDs exposing onboard sensor data, such approaches may enable fully self-contained, large-scale multi-user co-location without external tracking infrastructure.

\section{Conclusion}

We presented a robust and practical framework for enabling co-located multi-user VR, which aligns users in virtual space in a same physical immediate space. The novelty of our method consists in combining inside-out HMD tracking with occasional motion capture-based alignment. Unlike systems that rely either on one-time alignment or on continuous external tracking, our approach completely mitigates the latency introduced by continuous pose streaming which leads to grave issue of cybersickness. Instead, it preserves the responsiveness and fluidity of native HMD tracking while maintaining spatial consistency across users through sparse realignment events.

The reliability of the system has been validated through controlled experiments involving complex motion patterns known to challenge tracking stability. Notably, the virtual tracking space alignment for multi-user co-location was successful in all fist bump trials, demonstrating robust alignment even in dynamic and challenging interaction scenarios. 

We believe the system is ready to be scaled for testing to significantly larger environments and user groups of ten or more. Additionally, our next steps may include extending co-location to include physical objects, enabling richer interaction between users and their environment. In parallel, exploring map sharing through collaborative SLAM may allow a more self-reliant and streamlined system, ultimately reducing dependency on external infrastructure while maintaining precise multi-user alignment. 

\vspace{-0.75em}
\section*{Acknowledgments}
\vspace{-0.5em}

 funded by the Deutsche Forschungsgemeinschaft (DFG, German Research Foundation) under Germany’s Excellence Strategy – EXC 2117 – 422037984
\vspace{-1.0\baselineskip}

\bibliographystyle{IEEEtran}
\bibliography{main.bib}

\end{document}